\documentclass[3p, 12pt, times]{elsarticle}

\usepackage{graphicx}
\usepackage{caption}
\usepackage[labelfont=bf]{caption}
\usepackage{subcaption}
\usepackage{float}
\usepackage{amsmath}
\usepackage{amssymb}
\usepackage{zed-csp}
\usepackage{makecell}
\usepackage{multicol}
\usepackage{multirow}

\begin{document}

\begin{frontmatter}
\title {Towards a Formal Specification for Self-organized Shape Formation in Swarm Robotics}
\date{\vspace{-5ex}}
\author[add1]{Yasir R. Darr} 
\author[add1]{Muaz A. Niazi$^*$}
\address[add1]{Computer Science Department,\\ COMSATS Institute of Information Technology,\\ Islamabad, Pakistan \\
$^*$Corresponding author\\ muaz.niazi@gmail.com}


\begin{abstract}
The self-organization of robots for the formation of structures and shapes is a stimulating application of the swarm robotic system. It involves a large number of autonomous robots of heterogeneous behavior, coordination among them, and their interaction with the dynamic environment. This process of complex structure formation is considered as a complex system, which needs to be modeled by using any modeling approach. Although, formal specification approach along with other formal methods has been used to model the behavior of robots in a swarm. However, to the best of our knowledge, the formal specification approach has not been used to model the self-organization process in swarm robotic system for the shape formation. In this paper, we use a formal specification approach to model the shape formation task of swarm robots. We use Z (Zed) language of formal specification, which is a state-based language to model the states of the entities of the systems. We demonstrate the effectiveness of Z for the self-organized shape formation. The presented formal specification model gives the outlines for designing and implementing the swarm robotic system for the formation of complex shapes and structures. It also provides the foundation for modeling the complex shape formation process for swarm robotics using a multi-agent system in a simulation-based environment.
\end{abstract}
\begin{keyword}
Swarm robotics \sep Self-organization \sep Formal specification   \sep Complex system 

\end{keyword}

\end{frontmatter}

\section{Introduction}
\label{sec:intro}
A self-organization of robots is observed for the formation task of swarm robotics systems. This involves a number of robots interacting with each other as well as with the environment for achieving the global behavior and completing the task. A self-organized formation includes the formation of structures, shapes \cite{meng2013morphogenetic}, and generating patterns \cite{oh2017bio} 
etc, by a large number of small programmable robots. The robots in a swarm aggregate \cite{vardy2016aggregation} 
or segregate in 
order to form the desired structure. Predefined information of desired structure or shape is given to all the robots in a swarm or it is achieved through adaptability of robots, interacting with the environment.

The formation of complex structures and shapes by the robots in a swarm using self-organization process can be considered as a complex system \cite{chen2016describing}. 
It is because, a vast number of homogenous or heterogeneous robots interact and cooperate with each other in a decentralized and dynamic environment at a large-scale, which relates it to Complex adaptive systems (CAS) \cite{2012cognitive}. 
The realization of such complex scenario is problematic as well as the implementation and modeling of such complex systems is a challenging task.

Complex systems can be modeled using formal methods, involve the modeling of each entity of the system \cite{hall1990seven}. The decomposition of the complex system into subsystems allows modeling each subsystem formally. These individual formal models can be helpful in understanding the whole system, by analyzing the dynamic behavior of these models against each component of the system. In addition, this helps in reducing the complexity of the system as well as with pre-analysis, error or flaws can be addressed.

Previous studies show the application of the formal specification approach in many domains for modeling various CAS. The formal specification used for the analysis for researcher scholars of different domains \cite{hussain2014toward}. 
For modeling the wireless sensor network, formal specification framework has been presented in studies \cite{afzaal2016formal}. 
Complex systems using Agent-Based Modelling (ABM) has also been modeled using formal specifications, such as modeling AIDS spread process using agent-based modeling \cite{siddiqa2017novel}. Likewise in \cite{niazi2011novel}, 
an agent-based framework for wireless sensor network as the complex adaptive environment. In energy and power domain, smart grid as a complex system presented using formal specification \cite{akram2017formal}.
 Formal methods applied for prediction in swarm-based systems 
\cite{rouff2004properties}.
 Similarly, a formal specification used in the swarm robotics domain for analyzing the behavior of robots in a swarm \cite{massink2013use, li2016formal}. 
It has also been used for the verification of swarm robotics system \cite{dixon2012towards}.

However, to the best of our knowledge, the formal specification framework has not been used to model the self-organized shape formation process for the swarm robotics systems using ABM. Therefore, it is required to model such interesting application of swarm robotics by using formal specification framework along with ABM tool mentioned in previous studies \cite{niazi2009agent}. Modeling such swarm robotics system using a formal framework, gives better understanding such CAS. In addition, it allows analyzing the constraints and flaws exist in self-organized shape formation by swarm robots, in a design phase and increase adaptability and scalability of the system.

In this paper, we establish the formal specification framework for modeling the self-organized shape formation process in swarm robotics domain. We analyze the entities involved in the formation process for swarm robots and illustrate their formal specifications.

The remaining structure of the paper is as follows: First we present Background foundations, then in method section, we discuss the informal description of shape formation model and formal framework for self-organized shape formation. While at the end of this paper, conclusion and future work is mention. 

\section{Theoretical Foundation}
\label{sec:backgroud}
In this section, we present the basic background concepts needed to understand the presented formal specification model. formal methods.
\subsection{Swarm Robotics}
Swarm robotics approach is based on swarm intelligence
 a phenomenon that is inspired by the biological swarm in nature having social and collective behavior. A large number of robots in a swarm coordinates with each other through local interaction and simple rules, thus able to achieve global behavior, which is scalable, flexible, and robust \cite{brambilla2013swarm}.
  A swarm robotics is being used for addressing real-world tasks. These tasks include aggregation, navigation, path formation foraging, and flocking \cite{bayindir2016review}.

\subsection{Formal specification}
\label{sec:formspec}
The formal specification is a technique used for modeling different types of systems mathematically. A system developer can comment about the completeness of the model for all aspects needed in specific abstraction level. The one significant feature of formal specification is that through modeling, it can effectively represent the real-world scenarios at various abstract levels. A formal specification language known as ``Z", developed by Oxford University in 1970. It can easily be converted into executable program code \cite{woodcock1996using}. 
A specification using ``Z" comprises of sets and predicates from the mathematical notation \cite{bowen1996formal}. 
This specification language is being used for modeling large-scale real-world complex systems 
\cite{hall1990seven}. 
It is also used to relate the agent logic in MAS \cite{d2004understanding}.

\section{Method}
\label{sec:methodology}
In this section, we present the informal specification of the self-organized shape formation for the swarm robotics based complex system as well as the formal framework for modeling the CAS of self-organized shape formation for large-scale swarm robots.

\subsection{Informal specification}

Modeling the shape formation in swarm robotics systems involves the self-organization of robots in a large swarm. The robots in a swarm are small, programmable, and mobile entities, which move randomly in the environment. All the robots interact with each other as well as with the environment during the formation process. A swarm may comprise of homogeneous robots having same behavior to achieve collective behavior, or it has heterogeneous robots having different capabilities, functionalities, and roles for achieving the global outcome. The information related to the desired shape or structure can be given to all the robots as an image or predefined coordinates in the environment, before the start of the process. They can also achieve the shape formation using their adaptive behavior in interaction with the environment.\\
    
States of the robot changes during the formation process. Before the start of the process, all the robots are stationary and not moving. When the process starts, all the robots start moving randomly in the environment and have unlocalized state while the shape is in empty state. After converging to one of the locations inside the shape coordination, the robot stops moving and become stationary and set its state to localized while the shape is in the partially filled state. The formation process continues in this way until one of the three-condition meets. The formation process terminates if the shape is completely formed and no more moving robot left, i.e. all the robots are localized. For the second condition, the process stops if few robots join the shape and no moving robots left, while the shape is incomplete. The third condition for the self-organization process for shape formation to stop is that shape is complete form and there are some moving robots still waiting to join the shape.
    
\subsection{Formal specification for self-organized shape formation}

We present the formal descriptions of our self-organized shape formation model using z-language for formal specification. The purpose of this is to validate all the informal requirements of the system by using formal specification. In this study, we consider two entities as robot and shape. We use ``sets" along with operation or state ``schemas", for declaring user-defined entities such as robot and shape regarding shape formation task. 
There are two entities named as \textit{robot} and \textit{shape}, involved in the self-organized shape formation process of swarm robotics system. Table \ref{tbl:components} shows the overview summary of the entities, their states, and events which cause the transitions of states for each of the entity.

\begin{table}[H]
\caption{Shape formation entities, states, and events}
\label{tbl:components}
\centering
\begin{tabular}{| p{2.5cm} | p{4.5cm} | p{4.0cm} |}
\hline
\textbf{Entity} & \textbf{States} &	\textbf{Events}  \\\hline

\multirow{3}{*}{Robot} 
& Stationary
&Start move\\
&UnLocalize &
Fault occur \\
& Localize &
Join shape\\
         
\hline
\multirow{3}{*}{Shape} 
    & Empty&
Robot join\\

&Partial&
Swarm join\\

&Complete&\\

\hline

\end{tabular}
\end{table}

\subsubsection{Robot in a swarm}
\begin{figure}[H]
\begin{center}

\includegraphics[width = 7.0 cm, height = 7.0 cm]{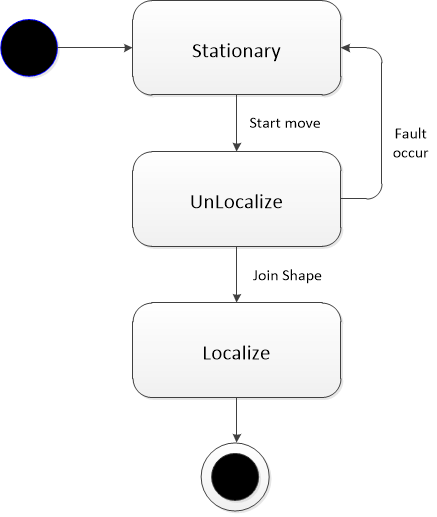}
\caption{The states of robot in a swarm.}
\label{fig:RobotState}

\end{center}
\end{figure}

First we discuss the specifications for the robot entity, its states, and events as shown in the figure \ref{fig:RobotState}. A robot is initially in the stationary state. After start move event, its state become unlocalize as it start moving randomly in the coordinate system. While a robot can again become stationary if any fault occur in it. When robot enters the desired shape coordinates and joins the already localized robots, it becomes localize. 

We define the ``max-xcor" and ``min-xcor" coordinate values for x coordinate as 32 and -32 respectively and similarly for the other two coordinates y and z as a ``world" or environment boundary for the robots.

\begin{axdef}
xcor, ycor, zcor : \num
\where
min$-$xcor=$-32$\\
max$-$xcor=$-32$\\
min$-$ycor=$-32$\\
max$-$ycor=$-32$\\
min$-$zcor=$-32$\\
max$-$zcor=$-32$
\end{axdef}	

\textit{Location} schema defines the environment area, where a robot can be placed randomly having  x,y,z coordinates in the cartesian coordinate system bounded between ``min-cor" and ``max-cor" values for coordinate parameters.

\begin{schema}{Location}
x,y,z : \num
\where
x\geq min$-$xcor\\
x \leq max$-$xcor\\
y \geq min$-$ycor\\
y \leq max$-$ycor\\
z \geq min$-$zcor\\
z \leq max$-$zcor
\end{schema}

The \textit{ROBOTSTATE} free type set defines a number of state a robot can have during the formation process. The states of robot consist of stationary, unlocalize, and localize. 
\begin{zed}
[ROBOTSTATE] == \{stationary, unLocalize, localize \}
\end{zed}

The \textit{Robot} schema contains ``robState" variable of the type \textit{ROBOTSTATE} to describe the current state of the robot.
\begin{schema}{Robot}
robState: ROBOTSTATE

\end{schema}

The ``robState" variable has ``stationary" value for robot state at the start as shown in the \textit{InitRobot} schema below. 

\begin{schema}{InitRobot}
Robot
\where
robState = stationary
\end{schema}

After the process start, robot starts moving randomly, so its state changes to ``unLocalize"" from ``stationary" as shown below in \textit{StartMove} schema. The change for robot state is represented by $\Delta$. 

\begin{schema}{StartMove}
\Delta Robot
\where
robState = stationary\\
robState' = unLocalize
\end{schema}

The \textit{FaultOccur} schema defines the change of state for robot which stops moving upon occurrence of any fault in itself and its state again changes to ``stationary".

\begin{schema}{FaultOccur}
\Delta Robot
\where
robState = unLocalize\\
robState' = stationary

\end{schema}

When a robot enters in the desired shape coordinates and stops adjacent to one of the localized robots, its state changes to ``localize" as describes in the below \textit{JoinShape} schema.\\

\begin{schema}{JoinShape}
\Delta Robot
\where
robState = unLocalize\\
robState' = localize
\end{schema}

We consider both perspectives of successful and error outcome regarding operational schemas for robot entity such as \textit{StartMove}, \textit{FaultOccur}, and  \textit{JoinShape} as shown in table \ref{tbl:freeType}.

\begin{table}[H]
\centering
\caption{Success and error outcomes of the schemas for \textit{robot} entity}
\label{tbl:freeType}

\begin{tabular}{| c | c | c |}
\hline
\textbf{Schema} & \textbf{Pre-condition for success} &\textbf{Condition for error} \\\hline

\multirow{2}{*}{StartMove} 
& Robot not moving
& Already moving:\\
&robState = stationary &
 robState = unLocalize \\        
\hline

\multirow{2}{*}{FaultOccur} 
& Robot moving
& Already faulted:\\
&robState = unLocalize &
 robState = stationary \\ 
\hline

\multirow{2}{*}{JoinShape} 
& Robot moving
& Already joined:\\
&robState = unLocalize &
 robState = localize\\
        
\hline

\end{tabular}
\end{table}

\subsubsection{Desire shape}
\begin{figure}[H]
\centering
\includegraphics[width = 7.0 cm, height = 7.0 cm]{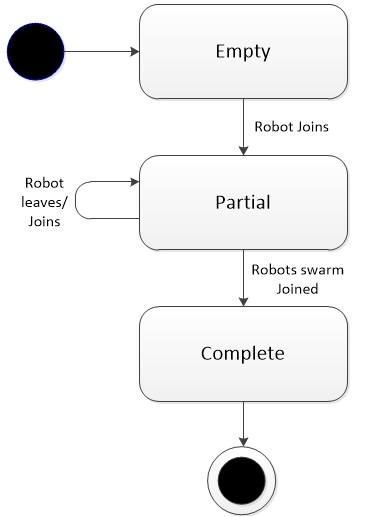}
\caption{The states of shape during formation process.}
\label{fig:ShapeState}
\end{figure}

Next we define the \textit{shape} entity, its states, and events for state transition. As figure \ref{fig:ShapeState} demonstrates the states of the shape which is empty at the start. While the other two states of the shape are ``partial" and ``complete".
At the start of the process shape is in ``empty" state. When robots start entering and joining the shape, the sate of the shape changes to ``partial". While robots in swarm become localize to completely form the shape, state of the shape changes to ``complete".

A free type set \textit{SHAPESTATE} defines the states of the desired shape during the formation process. It contains ``empty ", ``partial", and ``complete" values as the states of the shape.\\  

[SHAPESTATE] == \{ empty, partial, complete \}\\


We define the ``shapeDimension" variable to  evaluate the size and contour of the desired shape to be formed.  

\begin{axdef}
shapeDimention : \nat
\where
shapeDimention \geq 0
\end{axdef}

The schema \textit{shape} describes the shape to be formed in the coordinate system and shape dimension value obtained from the product of coordinates value of the shape. We define a variable ``sState" of the type \textit{SHAPESTATE} whose value can be ``empty", ``partial", and ``complete". A desired shape comprises of robots so we are including collection of robot schema as finite set of robots in declaration.

\begin{schema}{Shape}
sState:SHAPESTATE\\
robots:\finset Robot\\
p,q :\num
\where
p\geq min$-$xcor\\
p \leq max$-$xcor\\
q \geq min$-$ycor\\
q \leq max$-$ycor\\
shapeDimention = p*q\\
\end{schema}

At the start of the formation process, the shape is in empty state. That is, yet no robot has converged and joined the shape. As we are depicting that all robots are either in ``stationary" state or moving randomly in ``unLocalize" state. The schema \textit{InitShape} describes the state of shape ``empty" shown in ``sState" variable. 

\begin{schema}{InitShape}
Shape
\where
\forall rob : robots | rob .robState = unLocalize \\
sState = empty
\end{schema}

A robot moving randomly enters and joins the shape upon reaching next to one of the localized seed robot near origin inside the shape. A state of the robot given in ``robState" variable is ``localize" and the state of the shape is updated from ``empty" to ``partial" in
``sState" variable.

\begin{schema}{RobotJoins}
\Delta Shape

\where
sState = empty\\
\exists rob : robots |rob .robState = localize \\
sState' = partial
\end{schema}

While formation process continues for the shape,  and all the robots in swarm converge and become localized inside shape coordinates. A desired shape is completely formed by a number
of localized robots, a value for ``sState" variable is updated from ``partial" to
``complete".

\begin{schema}{RobotSwarmJoin}
\Delta Shape
\where
sState = partial\\

\forall rob : robots | rob .robState = localize \\
sState' = complete
\end{schema}

We also consider both the perspectives of successful and error outcome regarding operational schemas for shape entity as \textit{RobotJoin} and \textit{RobotSwarmJoin}.
As shown in table \ref{tbl:freeType2}.

\begin{table}[H]
\centering
\caption{Success and error outcomes of the schemas for \textit{shape} entity}
\label{tbl:freeType2}

\begin{tabular}{| c | c | c |}
\hline
\textbf{Schema} & \textbf{Pre-condition for success} &\textbf{Condition for error} \\\hline

\multirow{3}{*}{RobotJoins} 
& Robot moving and not joined:
& Already joined:\\
&robState = stationary 
& robState = unLocalize\\ 
& sState = empty & 
 sState = partial\\
         
\hline

\multirow{3}{*}{RobotSwarmJoin} 
& Some moving and some joined:
& Shape is complete:\\
&robState = unLocalize 
& robState = localize\\ 
& sState = partial & 
 sState = complete\\
\hline

\end{tabular}
\end{table}

\subsubsection{Schemas for Success and Error}

We define a free type definition \textit{REPORT} for success and error outcome for both \textit{robot} and \textit{shape} entity involved in the shape formation process. The elements of this free type are ``success", ``alreadyMoving", ``alreadyFaulted",  ``alreadyJoined",and ``alreadyCompleted".\\    
 
\begin{zed}
REPORT ::= success | alreadyMoving | alreadyFaulted | alreadyJoined | shapeCompleted
\end{zed}

\textit{Success} schema shows the success outcome for all operational schemas of \textit{Robot} and \textit{Shape} entity.  For the  robot entity, it shows that the robot successfully starts move. If fault occurs, it stops again, joins shape and become localize. Similarly for the shape entity, it shows normal process for shape formation from empty shape to partial after robots start joining and complete shape upon swarm joined shape. The report variable gives ``success" output value for all operational schemas.

\begin{schema}{Success}
report! : REPORT
\where
report!= success
\end{schema}

A robot is in ``unLocalize" state and not in ``stationary state". The schema for error outcome of \textit{StartMove} schema reports ``alreadyMoving" as output.  

\begin{schema}{AlreadyMoving}
\Xi Robot\\
report! : REPORT
\where
robState = unLocalize\\
report!= alreadyMoving
\end{schema}

If the robot is not moving due to some fault. It will not move and will be already in stationary state. The report output variable will give ``alreadyFaulted" value.

\begin{schema}{AlreadyFaulted}
\Xi Robot\\
report! : REPORT
\where
robState = stationary\\
report!= alreadyFaulted
\end{schema}

Robot after moving in unlocalize state converges and has joined the shape and is in localize state. So is the shape, the state of the shape is partial. Report variable shows ``alreadyJoined". 

\begin{schema}{AlreadyJoined}
\Xi Shape\\
report! : REPORT
\where
sState = partial\\
robState = localize \\
report!= alreadyJoined
\end{schema}

If robots in swarm joined the shape and have localize state as well as shape has complete state, then report output variable for error outcome schema shows `alreadyCompleted`" value. 

\begin{schema}{AlreadyCompleted}
\Xi Shape\\
report! : REPORT
\where
sState = complete\\
robState = localize \\
report!= alreadyComplete
\end{schema}

\subsubsection{Schema calculus}

The conjunction is used to combine two schemas and disjunction for combining alternative schemas. We combine \textit{StartMove} and \textit{success} along with alternative schema \textit{alreadyMoving} and define them in \textit{Move} schema. 

\begin{zed}
Move \defs (StartMove \land success) \lor alreadyMoving 
\end{zed}

For combining the success and error outcome of the \textit{Faultoccur} schema, we define \textit{Fault} schema which shows that robot is in normal working condition before the fault occurs or it can be faulted already, as given below. 
\begin{zed}
Fault \defs (FaultOccur \land success) \lor alreadyFaulted 
\end{zed}

We combine success outcome of \textit{JoinShape} schema for the shape formation process by using conjunction along with error outcome by using disjunction. We define \textit{Join} schema to show shape join process and is defined as:

\begin{zed}
Join \defs (JoinShape \land success) \lor alreadyJoined 
\end{zed}

We define a schema \textit{ShapePartial} to combine both success and error scenarion for the robots joining the shape for the formation. It shows the combination as  \textit{RobotJoins} and \textit{success} or \textit{alreadyJoined} 
\begin{zed}
ShapePartial \defs (RobotJoins \land success) \lor alreadyJoined 
\end{zed}

We define \textit{ShapeComplete} schema for combining the schemas involved in shape completion process. It is narrated as \textit{RobotSwarmJoin} and\textit{ success}, or \textit{alreadyComplete.} 

\begin{zed}
ShapeComplete \defs (RobotSwarmJoin \land success) \lor alreadyComplete 
\end{zed}

\section{Conclusion}
\label{sec:conclusion}
In this paper, we established the formal specification for the self-organized shape formation of swarm robotics system. There exist two entities in the system regarding shape formation process, i.e. a robot and the shape. We described every state and events for a state transition for each individual entity of the system. Further, we used a formal specification technique to model each entity involved in the shape formation process. This helps in better understanding the shape formation process by analyzing the behavior for each entity of the system. It depicted the usefulness of the formal specification technique for modeling such complex system. It also established the outlines for the development and implementation of the swarm robotics system to achieve the complex shape formation process. In future, we will use the formal specification
approach for modeling various other tasks such as path formation, navigation or
foraging, of the swarm robotics systems. 
\newpage
\textbf{References}
\bibliography{sample}

\begin{thebibliography}{10}

\bibitem{meng2013morphogenetic}
Yan Meng, Hongliang Guo, and Yaochu Jin.
\newblock A morphogenetic approach to flexible and robust shape formation for
  swarm robotic systems.
\newblock {\em Robotics and Autonomous Systems}, 61(1):25--38, 2013.

\bibitem{oh2017bio}
Hyondong Oh, Ataollah~Ramezan Shirazi, Chaoli Sun, and Yaochu Jin.
\newblock Bio-inspired self-organising multi-robot pattern formation: A review.
\newblock {\em Robotics and Autonomous Systems}, 91:83--100, 2017.

\bibitem{vardy2016aggregation}
Andrew Vardy.
\newblock Aggregation in robot swarms using odometry.
\newblock {\em Artificial Life and Robotics}, 21(4):443--450, 2016.

\bibitem{chen2016describing}
Chih-Chun Chen and Nathan Crilly.
\newblock Describing complex design practices with a cross-domain framework:
  learning from synthetic biology and swarm robotics.
\newblock {\em Research in Engineering Design}, 27(3):291--305, 2016.

\bibitem{2012cognitive}
Muaz~A Niazi and Amir Hussain.
\newblock {\em Cognitive agent-based computing-I: a unified framework for
  modeling complex adaptive systems using agent-based \& complex network-based
  methods}.
\newblock Springer Science \& Business Media, 2012.

\bibitem{hall1990seven}
Anthony Hall.
\newblock Seven myths of formal methods.
\newblock {\em IEEE software}, 7(5):11--19, 1990.

\bibitem{hussain2014toward}
Amir Hussain and Muaz Niazi.
\newblock Toward a formal, visual framework of emergent cognitive development
  of scholars.
\newblock {\em Cognitive Computation}, 6(1):113--124, 2014.

\bibitem{afzaal2016formal}
Hamra Afzaal and Nazir~Ahmad Zafar.
\newblock Formal analysis of subnet-based failure recovery algorithm in
  wireless sensor and actor and network.
\newblock {\em Complex Adaptive Systems Modeling}, 4(1):27, 2016.

\bibitem{siddiqa2017novel}
Amnah Siddiqa and Muaz~A Niazi.
\newblock A novel formal agent-based simulation modeling framework of an aids
  complex adaptive system.
\newblock {\em arXiv preprint arXiv:1708.02938}, 2017.

\bibitem{niazi2011novel}
Muaz~A Niazi and Amir Hussain.
\newblock A novel agent-based simulation framework for sensing in complex
  adaptive environments.
\newblock {\em IEEE Sensors Journal}, 11(2):404--412, 2011.

\bibitem{akram2017formal}
Waseem Akram and Muaz~A Niazi.
\newblock A formal specification framework for smart grid components.
\newblock {\em arXiv preprint arXiv:1711.09184}, 2017.

\bibitem{rouff2004properties}
Christopher Rouff, Amy Vanderbilt, Mike Hinchey, Walt Truszkowski, and James
  Rash.
\newblock Properties of a formal method for prediction of emergent behaviors in
  swarm-based systems.
\newblock In {\em Software Engineering and Formal Methods, 2004. SEFM 2004.
  Proceedings of the Second International Conference on}, pages 24--33. IEEE,
  2004.

\bibitem{massink2013use}
Mieke Massink, Manuele Brambilla, Diego Latella, Marco Dorigo, and Mauro
  Birattari.
\newblock On the use of bio-pepa for modelling and analysing collective
  behaviours in swarm robotics.
\newblock {\em Swarm Intelligence}, 7(2-3):201--228, 2013.

\bibitem{li2016formal}
Qin Li and Graeme Smith.
\newblock Formal development of multi-agent systems using maze.
\newblock {\em Science of Computer Programming}, 131:126--150, 2016.

\bibitem{dixon2012towards}
Clare Dixon, Alan~FT Winfield, Michael Fisher, and Chengxiu Zeng.
\newblock Towards temporal verification of swarm robotic systems.
\newblock {\em Robotics and Autonomous Systems}, 60(11):1429--1441, 2012.

\bibitem{niazi2009agent}
Muaz Niazi and Amir Hussain.
\newblock Agent-based tools for modeling and simulation of self-organization in
  peer-to-peer, ad hoc, and other complex networks.
\newblock {\em IEEE Communications Magazine}, 47(3), 2009.

\bibitem{barca2013swarm}
Jan~Carlo Barca and Y~Ahmet Sekercioglu.
\newblock Swarm robotics reviewed.
\newblock {\em Robotica}, 31(3):345--359, 2013.

\bibitem{brambilla2013swarm}
Manuele Brambilla, Eliseo Ferrante, Mauro Birattari, and Marco Dorigo.
\newblock Swarm robotics: a review from the swarm engineering perspective.
\newblock {\em Swarm Intelligence}, 7(1):1--41, 2013.

\bibitem{bayindir2016review}
Levent Bay{\i}nd{\i}r.
\newblock A review of swarm robotics tasks.
\newblock {\em Neurocomputing}, 172:292--321, 2016.

\bibitem{woodcock1996using}
Jim Woodcock and Jim Davies.
\newblock {\em Using Z: specification, refinement, and proof}, volume~39.
\newblock Prentice Hall Englewood Cliffs, 1996.

\bibitem{bowen1996formal}
Jonathan~Peter Bowen.
\newblock {\em Formal specification and documentation using Z: A case study
  approach}, volume~66.
\newblock International Thomson Computer Press London, 1996.

\bibitem{d2004understanding}
Mark d'Inverno, Michael Luck, and Michael~M Luck.
\newblock {\em Understanding agent systems}.
\newblock Springer Science \& Business Media, 2004.

\end{thebibliography}
\bibliographystyle{unsrt}

\end{document}